\documentclass[runningheads]{llncs}

\usepackage[T1]{fontenc}
\usepackage{graphicx,verbatim}
\usepackage{subcaption} 
\usepackage{amsmath,amssymb}
\usepackage{url}
\usepackage{makecell}
\usepackage{multicol,multirow}

\usepackage{xcolor}

\begin{document}
\title{Impact of Clinical Image Quality on Efficient Foundation Model Finetuning}
\titlerunning{Impact of Clinical Image Quality}
\author{Yucheng Tang\inst{1,2} \and 
Pawel Rajwa\inst{3} \and 
Alexander Ng\inst{3} \and
Yipei Wang\inst{1,2} \and
Wen Yan\inst{1,2} \and
Natasha Thorley\inst{4} \and
Aqua Asif\inst{3} \and
Clare Allen\inst{5} \and
Louise Dickinson\inst{5} \and
Francesco Giganti\inst{3,5} \and
Shonit Punwani\inst{4,5} \and
Daniel C. Alexander\inst{1,7} \and
Veeru Kasivisvanathan\inst{3,6} \and
Yipeng Hu\inst{1,2}
}
\authorrunning{Yucheng Tang, Pawel Rajwa, et al.}
%
\institute{
UCL Hawkes Insitute, University College London\and 
Dept. of Medical Physics \& Biomedical Engineering, University College London \and
Division of Surgery \& Interventional Science, University College London \and
Centre of Medical Imaging, University College London \and
Department of Radiology, UCLH NHS Foundation Trust \and
Department of Urology, UCLH NHS Foundation Trust \and
Department of Computer Science, University College London  \\
\email{yucheng.tang.24@ucl.ac.uk}
}



\maketitle              
\begin{abstract}

Foundation models in medical imaging have shown promising label efficiency, achieving high performance on downstream tasks using only a fraction of the annotated data otherwise required. In this study, we evaluate this potential in the context of prostate multiparametric MRI using ProFound, a recently developed domain-specific vision foundation model pretrained on large-scale prostate MRI datasets.
We investigate the impact of variable image quality on the label-efficient finetuning, by quantifying the generalisability of the finetuned models. We conduct a comprehensive set of experiments by systematically varying the ratios of high- and low-quality images in the finetuning and evaluation sets.
Our findings indicate that image quality distribution and its finetune-and-test mismatch significantly affect model performance. In particular:
a) Varying the ratio of high- to low-quality images between finetuning and test sets leads to notable differences in downstream performance; and
b) The presence of sufficient high-quality images in the finetuning set is critical for maintaining strong performance, whilst the importance of matched finetuning and testing distribution varies between different downstream tasks, such as automated radiology reporting and prostate cancer detection.
Importantly, experimental results also show that, although finetuning requires significantly less labeled data compared to training from scratch when the quality ratio is consistent, this label efficiency is not independent of the image quality distribution. For example, we show cases that, without sufficient high-quality images in finetuning, finetuned models may fail to outperform those without pretraining. 
This, in turn, highlights the importance of assessing (and potentially aligning) image quality distributions between finetuning and deployment, and the need for quality standards in finetuning data for specific downstream tasks. Using ProFound as a concrete example, this study demonstrates the value of quantifying image quality and its distribution in both finetuning and deployment to fully realise the data and compute efficiency benefits of foundation models.

\keywords{Foundation Model \and Image Quality \and Prostate MRI}

\end{abstract}
\section{Introduction}

Foundation models have shown strong transferability and generalization in medical image analysis~\cite{zhang2024challenges}, and have been applied across different anatomical regions and imaging modalities~\cite{khan2025comprehensive}, such as MultiTalent~\cite{ulrich2024multi} for multi-organ segmentation in abdominal computed tomography (CT), SAM-Med3D~\cite{wang2025sam}, which adapts the Segment Anything Model~\cite{zhang2024segment} to 3D medical images, SegVol~\cite{du2024segvol} for general CT segmentation tasks, and Adam~\cite{taher2024representing} for organ identification in chest X-rays and retinal images.
Based on these developments, some foundation model-based studies have emerged for prostate magnetic resonance imaging (MRI) related tasks, including cancer detection~\cite{wilson2024prostnfound}, modality classification~\cite{denner2025fine}, gland segmentation~\cite{kim2023empirical}, and cancer segmentation~\cite{zhang2025generalist}.
However, despite their success, the systematic examination of how prostate MRI quality affects the transferability and performance of foundation models in downstream tasks remains underexplored. In clinical practice, where variability in prostate MRI quality is common across different centers, scanners, and scanning protocols~\cite{woernle2024picture}, key questions arise: Can foundation models adapt effectively to data with varying quality levels? Does the transferability and performance of foundation models across image quality levels vary depending on the downstream tasks?

Previous work focused on the quality assessment of prostate MRIs, ranging from traditional feature-based approaches to deep learning-based quality classifiers~\cite{stkepien2022brief}. These studies examined the clinical impact of T2-weighted (T2W) image quality on diagnostic outcomes, including cancer detection~\cite{lin2024deep} and the evaluation of extraprostatic extension~\cite {lin2024deep2}. Subsequent studies applied the Prostate Imaging Quality Scoring System (PI-QUAL)~\cite{giganti2020prostate}, the international standard for evaluating prostate MRI quality, to multimodal prostate MRI scans~\cite{cheng2024impact} and explored the correlation between PI-QUAL scores and the detection rates of clinically significant prostate cancer at biopsy~\cite{brembilla2023impact}.

In this work, we analyze and reveal how prostate MRI quality affects the transferability and performance of finetuned foundation models across downstream tasks. Specifically, we utilize a publicly available, multi-task compatible prostate MRI foundation model, ProFound\footnote{\url{https://github.com/pipiwang/ProFound}}, and then fine-tune it using the PRIME dataset ~\cite{asif2023comparing}, a multi-institutional collection with PI-QUAL scores collected from 21 centers. We evaluated the finetuned model on four tasks:
1) Prostate imaging-reporting and data system (PI-RADS) score~\cite{turkbey2019prostate} 5-class classification (scores 1–5),
2) PI-RADS score binary classification ($\geq$3 vs. <3),
3) PI-RADS score binary classification ($\geq$4 vs. <4),
4) Gleason score~\cite{epstein2010update} binary classification ($\geq$3+4 vs. <3+4).

Our contributions are as follows: First, we curated the PRIME dataset from 21 centers, together with image quality PI-QUAL scoring. Second, after preprocessing of the data, with the image quality labels, radiological and histopathological disease labels, we finetuned the ProFound model using the preprocessed PRIME dataset across different downstream tasks. Furthermore, we compared and analyzed the generalisability of this independent data set by varying the ratios of high-quality and low-quality images used in both finetuning and testing, thus concluding the importance of quantifying image quality to the label efficiency due to foundation models.

\section{Method}

We utilize the ProFound-alpha model, a recently developed and released foundation model for prostate multiparametric MRI that was pretrained on large-scale datasets including PI-CAI~\cite{saha2023artificial}, Prostate-X~\cite{saha2023artificial}, as well as on other private datasets. 
It adopts a ConvNeXt V2 Tiny~\cite{woo2023convnext} backbone pretrained via masked autoencoding (MAE)~\cite{he2022masked} to learn generalizable imaging features. Model architecture and pretraining details are available in the official repository\footnotemark[9].
In addition, the ProFound repository provides finetuning code for different downstream tasks, which we used for finetuning in this study with the same hyperparameter configurations unless specified. The data preprocessing and further finetuning details are described in Section 2.1 and Section 2.2, respectively.

\subsection{Data Preprocessing}
The private PRIME dataset~\cite{asif2023comparing} contains 483 cases of prostate MRI with PI-QUAL scores assigned by expert radiologists (v2~\cite{de2024pi}), including 452 cases of high quality (PI-QUAL $\geq$ 4, possible to rule out all significant lesions) and 31 cases of low quality (PI-QUAL $<$ 4, not possible to rule in all significant lesions). 
We acknowledge that the dataset is imbalanced and relatively small, particularly for low-quality scans; however, this reflects the natural prevalence of high-quality imaging in clinical workflows.
For the PRIME dataset used in finetuning, we applied standardized preprocessing to ensure spatial consistency across multicenter scans. First, from each case, we selected three image sequences: (1) the diffusion-weighted imaging (DWI) sequence with the highest b-value (typically b=2000), (2) the axial T2W image with the highest in-plane resolution, and (3) the corresponding apparent diffusion coefficient (ADC) map. The transverse T2W image was used as the spatial reference for subsequent imaging coordinate alignment and resampling.
We then aligned the DWI and ADC maps to the T2W image reference using an inter-sequence coordinate transformation with SimpleITK library~\cite{lowekamp2013design}, ensuring consistent image origin, orientation, and direction across modalities. After alignment, images from the three modalities were resampled to a uniform voxel spacing of $0.5 \times 0.5 \times 1.0 \text{mm}^3$. Finally, each resampled volume was center-cropped and symmetrically zero-padded to a fixed spatial size of $224 \times 224 \times 64$ voxels to ensure consistent input sizes consistent with the pretrained ProFound model.

\subsection{Foundation Model Finetuning for Downstream Tasks}

Following the data preprocessing, we finetuned the ProFound foundation model on the processed data. The finetuning tasks are formulated as binary or multi-class classification problems to predict clinically relevant categories. Specifically, we address four downstream tasks: 1) multi-class classification of the PI-RADS scores into five categories corresponding to scores 1 through 5, aiming to stratify prostate lesions by their likelihood of clinically significant cancer; 2) binary classification of the PI-RADS scores, distinguishing cases with scores $\geq$ 3 vs. < 3, which is generally considered the threshold for suspicious lesions warranting further investigation; 3) binary classification of the PI-RADS scores, distinguishing cases with scores $\geq$ 4 vs. < 4, which is considered the threshold for lesions at greater risk; and 4) binary classification of the Gleason scores, separating $\geq$ 3+4 from < 3+4, which helps differentiate more aggressive prostate cancers that often demand active treatment from less aggressive cases.

Specifically, for each finetuning setting, a classification model is constructed by attaching a fully connected classification head, composed of two linear layers with ReLU activation, to the pretrained ConvNeXt V2 Tiny backbone. During finetuning, all model parameters are trainable. We apply layer-wise learning rate decay to the backbone, assigning smaller learning rates to layers closer to the input and higher rates to deeper layers. For supervision, we adopt the cross entropy loss as the loss function in finetuning stage, which is defined as in Eq.~\ref{FLoss}.
\begin{equation}
\label{FLoss}
\mathcal{L}_{fine} = -\frac{1}{N}\sum_{n=1}^{N}\sum_{c=1}^C y_{n,c} \log(\frac{\exp(\hat{y}_{n,c})}{\sum_{k=1}^C\exp(\hat{y}_{n,k})})
\end{equation}
where $N$ is the number of samples, $C$ is the number of classes, $\hat{y}_{n,c}$ is the predicted logit for the $c$-th class of the $n$-th sample, and $y_{n,c}$ is the one-hot encoded ground truth label.
This finetuning scheme enables the foundation model to adapt to the specific data and task while leveraging the pretrained spatial feature representations learned from large-scale data.

\section{Experiments and Results}

\textbf{Implementation Datails.}
The ProFound model was pretrained for 800 epochs under a masked autoencoding framework with a mask ratio of 0.6, and we used the checkpoint of the ProFound model at the 800$^{th}$ epoch for downstream finetuning.
The finetuning process was conducted using PyTorch~\cite{paszke2019pytorch}, and was run for 100 epochs on the preprocessed PRIME dataset with a batch size of 8, an input size of $224 \times 224 \times 64$, and a maximum learning rate of $1 \times 10^{-4}$ assigned to the deepest layer group.
We adopted a layer-wise learning rate decay strategy with a decay factor of $0.9$ applied to groups of three consecutive layers, where each shallower group receives a learning rate $0.9$ times that of the next deeper group. All experiments were conducted on NVIDIA Quadro GV100 GPUs, each with 24 GB of memory.

\textbf{Finetuning Data Settings.}
To evaluate the transferability and performance of the finetuned foundation model under different data quality scenarios, we adopted three data settings for finetuning: (1) Mixed-finetuned: using both high and low-quality data (180 high-quality and 16 low-quality cases), followed by affine augmentation (detailed in Section 3.1) to produce 540 cases in each group. The resulting model is referred to as the 'mixed-finetuned model' in the following text; (2) HQ-finetuned: using 180 high-quality cases augmented to 1080 cases. The resulting model is referred to as the 'HQ-finetuned model' in the following text; and (3) LQ-finetuned: using 16 low-quality cases augmented to 1080 cases. The resulting model is referred to as the 'LQ-finetuned model' in the following text. 
The affine transformation was applied jointly to all three modalities (T2W image, ADC, and DWI) of each patient case, using random rotation (up to $15^\circ$) and translation (up to 0.15 mm) while preserving spatial alignment.

\textbf{Testing Data Settings.}
For testing, we fixed the size of the test set to 60 cases and varied the sampling ratio of high-quality to low-quality data at five levels: 14:1, 11:1, 8:1, 5:1, and 2:1. For each ratio, samples were drawn from the remaining cases not used in training. We performed five independent samplings per ratio and reported the average and standard deviation of evaluation metrics.

\begin{figure}[t]
    \centering
    \begin{subfigure}[t]{0.48\textwidth}
        \centering
        \includegraphics[width=\linewidth,height=4cm]{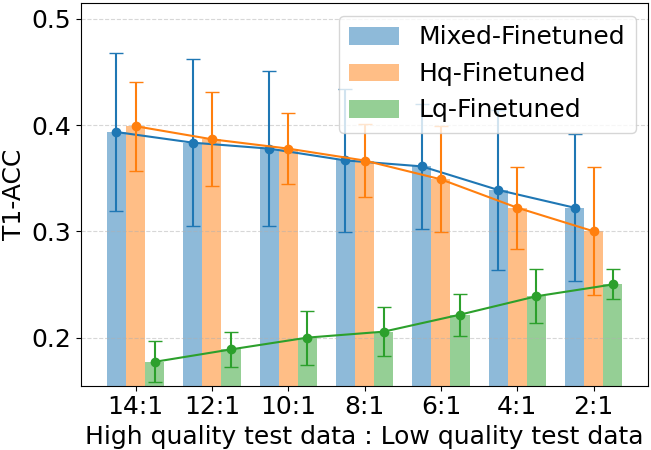}
        \caption{PI-RADS 5-class classification}
        \label{PI-RADS_5_acc}
    \end{subfigure}
    \quad
    \begin{subfigure}[t]{0.47\textwidth}
        \centering
        \includegraphics[width=\linewidth,height=4cm]{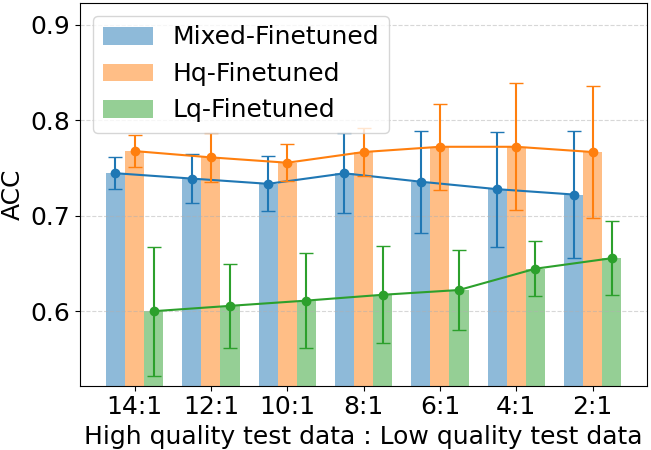}
        \caption{PI-RADS binary classification ($\geq$3 vs. <3)}
        \label{PI-RADS_2_3_acc}
    \end{subfigure}
    \begin{subfigure}[t]{0.48\textwidth}
        \centering
        \includegraphics[width=\linewidth,height=4cm]{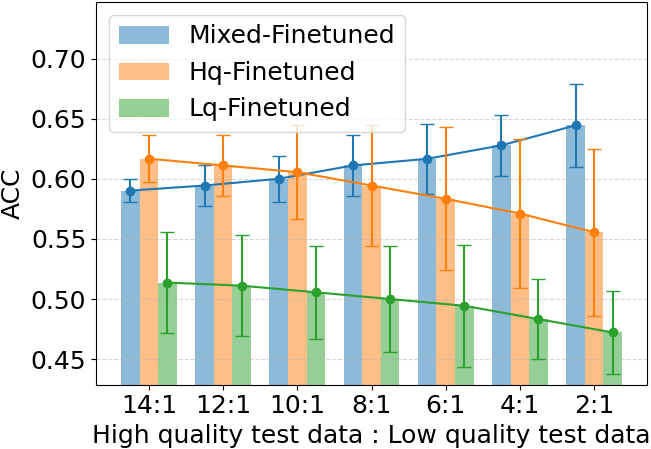}
        \caption{PI-RADS binary classification ($\geq$4 vs. <4)}
        \label{PI-RADS_2_4_acc}
    \end{subfigure}
    \quad
    \begin{subfigure}[t]{0.48\textwidth}
        \centering
        \includegraphics[width=\linewidth,height=4cm]{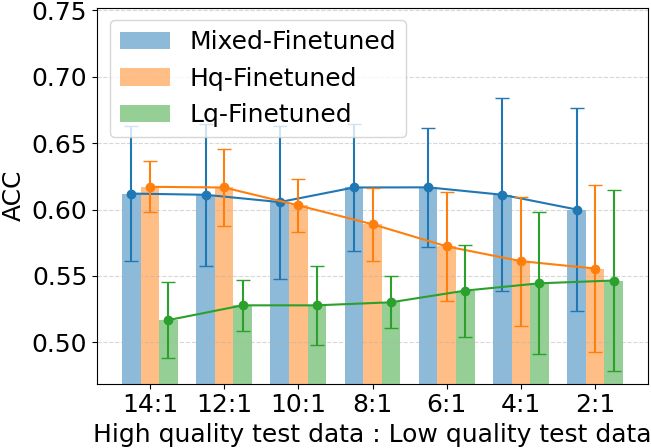}
        \caption{Gleason score binary classification ($\geq$3+4 vs. <3+4)}
        \label{Gleason_2_acc}
    \end{subfigure}

    \caption{Experimental results for the four downstream tasks. The evaluation metric is accuracy (and top 1 accuracy for 5-class classification). In the figures, the blue bars, line, and error bars represent the test accuracy (with standard deviation) of the foundation model finetuned on both high- and low-quality data. The orange elements represent the model finetuned on high-quality data only, and the green ones correspond to the model finetuned on low-quality data only.} 
    \label{ACC}
\end{figure}

\textbf{Evaluation Metrics.}
For binary classification tasks aimed at identifying clinically significant lesions (e.g., PI-RADS $\geq$ 3 vs. $<$ 3 or $\geq$ 4 vs. $<$ 4, Gleason score $\geq$ 3+4 vs. $<$ 3+4), we evaluated model performance using accuracy (ACC) and area under the ROC curve (AUC). For multi-class classification tasks (e.g., PI-RADS 1–5), we reported top-1 accuracy (T1-ACC) and AUC.

\begin{figure}[t]
    \centering
    \begin{subfigure}[t]{0.48\textwidth}
        \centering
        \includegraphics[width=\linewidth,height=4cm]{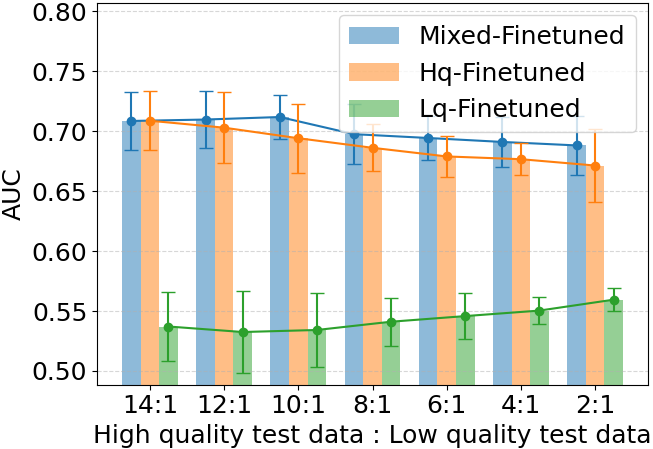}
        \caption{PI-RADS 5-class classification}
        \label{PI-RADS_5_auc}
    \end{subfigure}
    \quad
    \begin{subfigure}[t]{0.48\textwidth}
        \centering
        \includegraphics[width=\linewidth,height=4cm]{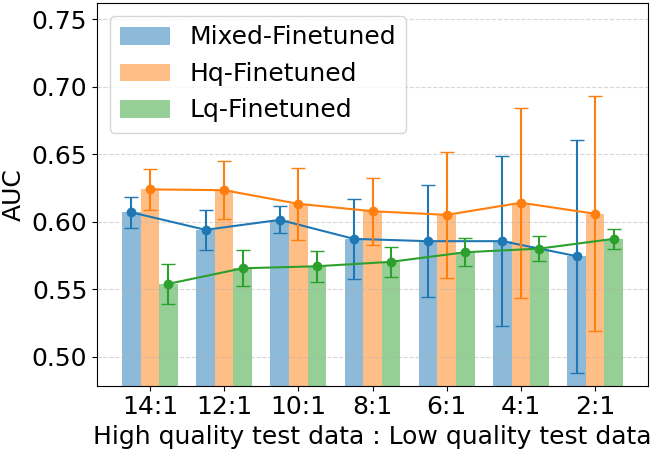}
        \caption{PI-RADS binary classification ($\geq$3 vs. <3)}
        \label{PI-RADS_2_3_auc}
    \end{subfigure}
    \begin{subfigure}[t]{0.48\textwidth}
        \centering
        \includegraphics[width=\linewidth,height=4cm]{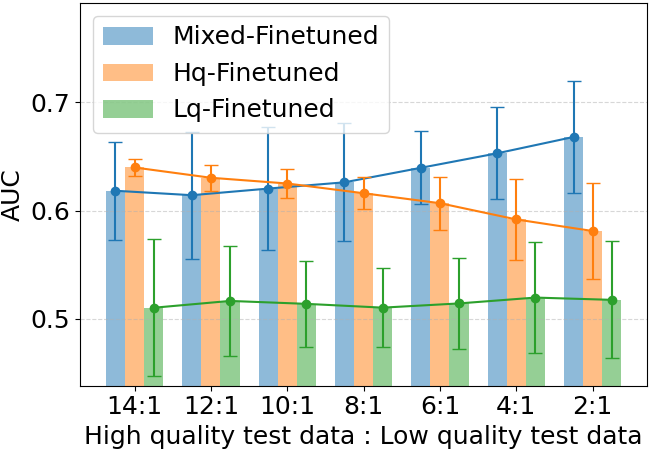}
        \caption{PI-RADS binary classification ($\geq$4 vs. <4)}
        \label{PI-RADS_2_4_auc}
    \end{subfigure}
    \quad
    \begin{subfigure}[t]{0.48\textwidth}
        \centering
        \includegraphics[width=\linewidth,height=4cm]{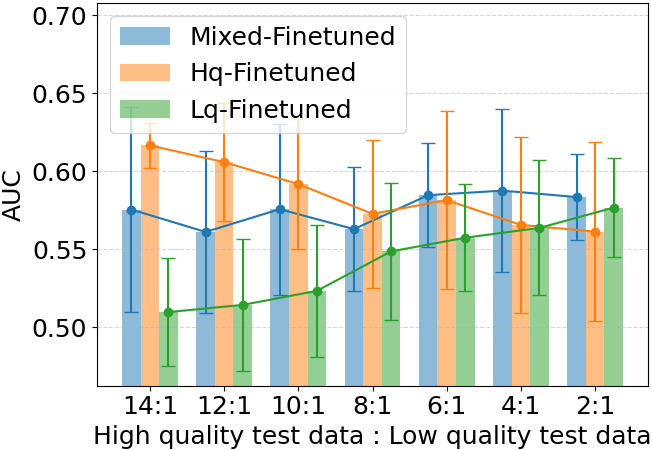}
        \caption{Gleason score binary classification ($\geq$3+4 vs. <3+4)}
        \label{Gleason_2_auc}
    \end{subfigure}

    \caption{Experimental results for the four downstream tasks. The evaluation metric is AUC. Color and marker conventions are the same as in Figure~\ref{ACC}.} 
    \label{AUC}
\end{figure}

\textbf{PI-RADS score 5-class classification.} 
As shown in Figure~\ref{ACC} (a), and Figure~\ref{AUC} (a), both the mixed-finetuned model and the HQ-finetuned model exhibit a decreasing performance across all three metrics as the proportion of low-quality data increases in the test set, with the HQ-finetuned model showing a significantly steeper decline due to the difference between the finetuning and testing data distributions.
In contrast, the LQ-finetuned model demonstrates improved T1-ACC under these conditions, due to better alignment between its finetuning and testing data distributions.
Despite these trends, the mixed-finetuned model consistently outperforms the LQ-finetuned model across all test settings, with statistical significance confirmed by paired t-tests (p < 0.05). Therefore, finetuning with a combination of high- and low-quality data is recommended to ensure better performance stability across variable test conditions.

\textbf{PI-RADS score binary classification ($\geq$ 3 vs. <3).}
As shown in Figure~\ref{ACC} (b), and Figure~\ref{AUC} (b), the mixed-finetuned model exhibits performance degradation as the proportion of low-quality data increases in the test set, while the LQ-finetuned model shows improved performance as the test set becomes increasingly dominated by low-quality data. These trends are similar to those observed in the PI-RADS 5-class classification task. Notably, the HQ-finetuned model maintains stable performance even as the proportion of low-quality data increases in the test set. 
Furthermore, in this task, the mixed-finetuned model only significantly outperforms the LQ-finetuned model when the test set contains a high-to-low quality ratio of at least 8:1, with statistical significance confirmed by paired t-tests (p < 0.05). As the test distribution becomes dominated by low-quality samples, the performance of finetuning on mixed data decreases quickly. In contrast, the HQ-finetuned model consistently outperforms the LQ-finetuned model across all test settings, with statistical significance (p < 0.05), and even surpasses the mixed-finetuned model under most conditions. This suggests that incorporating low-quality data during finetuning compromises the model's transferability and generalization ability, and that high-quality data alone provides more stable and transferable feature representations for the PI-RADS binary classification task ($\geq$ 3 vs. < 3). 
Therefore, it is preferable to finetune the model using high-quality data.

\textbf{PI-RADS score binary classification ($\geq$ 4 vs. <4).}
The results of the PI-RADS binary classification task ($\geq$ 4 vs. < 4) are shown in Figure~\ref{ACC} (c), and Figure~\ref{AUC} (c). Surprisingly, unlike other tasks, the mixed-finetuned model shows improved generalization as more low-quality data was added to the test set. In contrast, the LQ-finetuned model achieves very poor test metrics, indicating that it fails to generalize to test sets that contain mostly high-quality data. The HQ-finetuned model also shows performance degradation as the proportion of low-quality data increases due to the difference between finetuning and testing data distributions.
Moreover, the mixed-finetuned model significantly outperforms the LQ-finetuned model across all test settings, and also outperforms the HQ-finetuned model when the high-to-low quality ratio is 2:1, with paired t-tests confirming statistical significance (p < 0.05). This suggests that, for the PI-RADS binary classification task ($\geq$ 4 vs. < 4), it is important to include low-quality data along with high-quality data during training to cover the full range of image variability and improve model robustness.

\textbf{Gleason score binary classification ($\geq$ 3+4 vs. <3+4).} 
The results of the Gleason score binary classification task ($\geq$ 3+4 vs. < 3+4) are shown in Figure~\ref{ACC} (d) and Figure~\ref{AUC} (d). The model finetuned on the mixed data demonstrates relatively stable generalization performance, showing limited sensitivity to increasing proportions of low-quality test data. In contrast, the HQ-finetuned and LQ-finetuned models perform well only within their respective familiar testing data distributions, but struggle to generalize across quality levels. Specifically, only when the high-to-low quality ratio in the test set exceeds 10:1, the HQ-finetuned model significantly outperforms the LQ-finetuned model, with statistical significance confirmed by paired t-tests (p < 0.05).
Moreover, this task shows higher variance in performance compared to the others, suggesting less consistent predictions, due to the inherent subjectivity and complexity of Gleason grading. These findings indicate that, for the Gleason score binary classification task, both high- and low-quality finetuning data are necessary. While each quality-specific model performs well in its own testing data's quality level, their limited transferability highlights the need for finetuning data that spans the full levels of image quality to improve robustness.

\textbf{Comparisons with models without pretrained weights.}
To investigate the transferability of the foundation model, we conducted experiments by training models from scratch, using randomly initialized weights but keeping the same model architecture and configurations as finetuning. Specifically, we trained models for the four tasks using the same datasets as in the finetuning setup, which demonstrated satisfactory performance in the experiments before.
Experimental results demonstrate that models trained from scratch converge more slowly, typically requiring 15 to 20 epochs to reach convergence, in contrast to 5 to 10 epochs for finetuning models. 
Furthermore, training from scratch using mixed and high-quality-only datasets led to performance drops ranging from 7.72$\%$ to 16.77$\%$ in ACC and 2.63$\%$ to 9.38$\%$ in AUC across different tasks, training settings, and high-to-low quality test ratios (as used in the finetuning setup).
These findings highlight the superior transferability of the foundation model, showing that its pretrained knowledge offers a strong initialization that not only speeds up convergence but also leads to consistently better performance across multiple classification tasks and varying data quality conditions. However, the results that support label efficiency are conditioned on consistent image quality ratios between models with and without pretraining. When comparing models with different imaging quality ratios, such gains in performance (and/or label efficiency) depend on the image quality distribution. For example, as more low-quality images were included in the finetuning set, the average accuracy gain from pretraining (across different tasks and high-to-low quality test ratios) dropped from +12.55$\%$ with high-quality-only data to +8.34$\%$ with mixed data, and became negative (–0.31$\%$) when using only low-quality data.

\section{Conclusion and Discussion}
In this work, we reveal the impact of prostate MRI quality on finetuning a foundation model. Our results show that models finetuned only on low-quality data consistently perform worse across all tasks as such data leads to irreversible information loss and artifact learning, degrading feature representations. This, in turn, disrupts clinical workflows and increases the risk of missed diagnoses and unnecessary biopsies.
While including high-quality data in finetuning generally improves performance gain due to pretraining. 
Moreover, such an impact from image quality to downstream performance is also task-specific. 
Among them, Gleason score binary classification task suffers the most due to its reliance on fine microstructural features, whereas PI-RADS ($\geq$4 vs. $<$4) classification is comparatively robust, benefiting from more resilient texture-based cues. Adaptive fine-tuning may offer a potential solution to such task-specific challenges.
Through the example of ProFound, our findings highlight that rigorous quantification of image quality and its distribution is critical to unlocking the full potential of foundation models in terms of data and computational efficiency.
Future work will benefit from expanding both the task diversity (e.g., segmentation or report generation) and model tuning strategies (e.g., adapter-based fine-tuning or quality-aware sampling).

\subsubsection{Disclosure of Interests.} The authors have no competing interests to declare that are relevant to the content of this article.


%
%
%
\bibliographystyle{splncs04}
\bibliography{bib.bib}

\end{document}